\documentclass{article}

    \usepackage[preprint]{neurips_2020}

\usepackage[utf8]{inputenc} 
\usepackage[T1]{fontenc}    
\usepackage{hyperref}       
\usepackage{url}            
\usepackage{booktabs}       
\usepackage{amsfonts}       
\usepackage{nicefrac}       
\usepackage{microtype}      
\usepackage{algorithm}
\usepackage[noend]{algpseudocode}
\usepackage{subcaption} 
\usepackage{graphicx}
\usepackage{amsmath}
\usepackage{bbm}
\usepackage{wrapfig}

\makeatletter
\algrenewcommand\ALG@beginalgorithmic{\footnotesize}
\makeatother

\setcitestyle{authoryear}

\title{From proprioception to long-horizon planning in novel environments: A hierarchical RL model}

\author{%
  Nishad Gothoskar \quad Miguel L\'azaro-Gredilla \quad Dileep George\\
  Vicarious AI\\
  \texttt{\{nishad,miguel,dileep\}@vicarious.com}
}

\begin{document}

\maketitle

\begin{abstract}
For an intelligent agent to flexibly and efficiently operate in complex environments, they must be able to reason at multiple levels of temporal, spatial, and conceptual abstraction. At the lower levels, the agent must interpret their proprioceptive inputs and control their muscles, and at the higher levels, the agent must select goals and plan how they will achieve those goals. It is clear that each of these types of reasoning is amenable to different types of representations, algorithms, and inputs. In this work, we introduce a simple, three-level hierarchical architecture that reflects these distinctions. The low-level controller operates on the continuous proprioceptive inputs, using model-free learning to acquire useful behaviors. These in turn induce a set of mid-level dynamics, which are learned by the mid-level controller and used for model-predictive control, to select a behavior to activate at each timestep. The high-level controller leverages a discrete, graph representation for goal selection and path planning to specify targets for the mid-level controller. We apply our method to a series of navigation tasks in the Mujoco Ant environment, consistently demonstrating significant improvements in sample-efficiency compared to prior model-free, model-based, and hierarchical RL methods. Finally, as an illustrative example of the advantages of our architecture, we apply our method to a complex maze environment that requires efficient exploration and long-horizon planning.
\end{abstract}

\section{Introduction}

A fundamental feature of any intelligent system is the ability to hierarchically organize information. Hierarchies enable abstraction and compositionality, both of which are important for efficient learning and inference. In the context of learning for control, abstraction of control signals allows low-level systems to handle actuation of specific muscles while higher-level systems can handle macro-level decision making. Compositionality allows us to construct more complex skills from simpler primitive behaviors, enabling reuse and therefore reducing the time needed to learn.

Another important advantage of hierarchy is that we can apply different representations and algorithms that better align with the type of reasoning needed at each level. Planning and reinforcement learning constitute two very different algorithms that operate on very different representations, however they are equally important. Planning operates on graphs, which are a useful way to represent the world, and allow us to solve long-horizon tasks. However, in continuous control settings, graphs and planning are not always feasible to use due to high-dimensional state spaces or stochasticity of dynamics. In this regime, deep reinforcement learning (RL) has shown great success through the use of rewards to learn policies.

But while deep RL has enabled agents to perform a variety of complex tasks, from game playing \citep{play} to control \citep{control}, this often comes at the cost of enormous amounts of training data. Simulation has helped ease this burden, along with methods that improve the Sim-to-Real transfer. However, if we attempted to apply these methods directly on a real robot they would often require many days of continuous training to succeed. Our goal is to leverage the advantages of both reinforcement learning and planning to build a system that can be trained with relatively few samples and is realizable on a real robot.

To that end, in this work we introduce a simple, hierarchical architecture that bridges model-free RL, model-based RL, and planning. We apply each of these in the regimes where they are most successful: (1) model-free RL at a low level to learn short-horizon behaviors, (2) model-based RL at an intermediate level where the dynamics of the action space are predictable (3) and planning at the high level where we can reliably move between states of the state space. By using these methods in a hierarchical manner, each level produces an abstraction that simplifies the control problem for the next level.

\section{Architecture}
Our architecture is a three-level hierarchy consisting of the \textit{behavior library}, the \textit{model learner}, and the \textit{planning system}, in bottom-up order. Each of these levels is described in detail in Section \ref{sec:lowlevel}, Section \ref{sec:midlevel}, Section \ref{sec:toplevel}, respectively. In Section \ref{sec:proprioception}, we discuss the assumptions we make about the agent's understanding of the state it observes. We present our architecture in the context of an Ant navigation task since this is our benchmark of choice \citep{gym}.

\subsection{Proprioception, External State, and Dimensions of Interest}
\label{sec:proprioception}

Proprioception is the awareness of the body's position and movement. In the case of an Ant, this would consist of the perception of the configuration of its joints and corresponding velocities. It would also include the forces sensed by the ant. These proprioceptive inputs are a subset of those provided as state in the standard OpenAI Gym Ant  environment \citep{gym}. The non-proprioceptive state dimensions are referred to as the external state i.e., those which cannot be directly sensed or affected by actuation of the agent's joints. In the case of the Ant, these are the x-y-z coordinates of its body in the environment as well as its orientation. In addition to the separation of proprioceptive and external state, we also assume the agent has a sense of which dimensions of its input it would like to actuate\footnote{This is a mild assumption also present in standard RL: when a 2D navigation reward is chosen, it is typically based on the x-y position, which implicitly injects the knowledge of which the dimensions of interest  are.}, as second-order effects of actuating its joints. In the case of Ant navigation, we assign the x-y coordinates as the dimensions of interest. This understanding is important in limiting the inputs received at higher levels in the hierarchy, since at the level at which we plan, we likely do not need access to the specific configuration of the joints and other proprioceptive inputs. 

We follow previous work on hierarchical RL that split inputs into proprioceptive and external inputs \citep{hrl}. It is not unreasonable to assume an agent would be able to identify which dimensions are proprioceptive. "Flat" deep reinforcement learning has been able to handle high dimensional observations and learn a selectivity for certain inputs. As such, it has not been necessary to limit the number of inputs. So, this categorization of state dimensions is not a standard practice in reinforcement learning literature.

For notation, we use $s^l$, $s^m$, and $s^h$ to refer to the state inputs used by each of the low, mid, and high levels. $s^l$ contains proprioceptive inputs, $s^m$ contains the external state, and $s^h$ contains the dimensions of interest.

\subsection{Low-Level: Behavior Library}
\label{sec:lowlevel}

Since the behavior library is the lowest level module in the architecture, it is responsible for the direct interaction with the environment, i.e., receiving the state as input and computing actions (motor commands). Standard model-free RL methods learn a single policy, often in the form of a neural network, that takes the state as input and computes actions. While there are many tasks that can be solved entirely by a single policy, learned in a model-free manner, at some level of task complexity, this will not be possible (at least not with a reasonable number of samples). In our work, we use \textit{behavior} to refer to a single policy. However, a single behavior is not responsible for solving the entire task on its own. Rather we seek to learn a set of behaviors, referred to as the \textit{behavior library}, that can actuate over the dimensions of interest and can be used in coordination  to solve the task.


We seek to learn behaviors that can create a consistent change to the external state $s^m$. We use the simple reward function below, that is parametrized by the vector $v$, which denotes the desired change of the mid-level state. We select a different $v$ to train each behavior. These could be chosen randomly, but to limit the number of behaviors needed, we select orthogonal vectors that span the dimensions of interest. We learn the behaviors using Twin Delayed DDPG \citep{td3}. Note, the behavior reward is only dependent on the external state $s^m$.
\begin{equation}
R_v (s_t,a_t,s_{t+1}) = 1 - \| (s^{m}_{t+1} - s^{m}_t) - v \|_1 
\end{equation}

This is a very simplistic reward function, that is clearly biased towards moving the state in a direction. We could easily substitute this with another unsupervised skill learning method like DIAYN, DADS, or HRL-EP3 \citep{diayn,dads,hrl}. In Section \ref{sec:learnthebehaviorlibrary} of the experiments, we discuss this issue in more detail in the context of the number of samples needed for each of these skill discovery methods.

In summary, the behavior library is a set of learned policies that serve as a useful abstraction for higher levels in the hierarchy to use, eliminating the need to directly interact with the environment.

\subsection{Mid-Level: Model Learner}
\label{sec:midlevel}

The mid-level module is responsible for learning how to use the behavior library to reach arbitrary goals. At each timestep, this module selects a behavior from the behavior library to execute. We use a model-based method that first learns the dynamics of each behavior and then uses these dynamics models in an MPC routine.

\subsubsection{Behavior Dynamics}

First, for each behavior $\pi_i$ contained in the behavior library $ \{ \pi_1, \pi_2, ... \pi_N \} $, we learn a corresponding dynamics model $f^i$. The parameters of these dynamics models are optimized according to the following objective:

\begin{equation}
\min_\theta  \mathbb{E}_{s_1, s_2,...,s_T \sim \pi_i} \Big[ \sum_{t=1}^{T-L} \| s^m_{t+L} - (s^m_t + f^i_\theta(s^m_t)) \|_2^2 \Big]
\end{equation}

The data used to learn this model is collected by executing the behavior and collecting sequences of observations. We sidestep the need to collect additional samples by reusing the samples from the final epochs of training for each behavior, as these are representative of the behavior's dynamics.

As is often done in dynamics learning, we are predicting $\Delta s^m$. However, our dynamics models are not predicting a one-timestep change, instead we have introduced a time scale parameter $L$ and we predict the change over $L$ steps of executing the behavior. In our case, since each behavior is trained to simply move linearly in the dimensions of interest, they are predictable at longer timescales. This is still applicable if we substituted our behavior learning with a method like DADS, since their learning objective includes a term that rewards negative entropy of the dynamics i.e. predictability \citep{dads}. This time scale parameter also provides temporal abstraction because a behavior at one step might be unpredictable but at longer time scales makes a consistent change to the state.

\subsubsection{Model-Predictive Control}

Now, using these dynamics models, we want to select a behavior to activate at each timestep. We do this online using Model-Predictive Control (MPC) \citep{mpc}. The goal of MPC, in general, is to optimize a sequence of actions to minimize an error with respect to a target state, execute the first action in this sequence, and then replan at the next timestep. In our setting, the sequence of actions is a sequence of behaviors to execute.

We optimize over a sequence of $H$ behaviors, where $H$ is the horizon. If there are $N$ behaviors, indexed $1,...,N$, this sequence is an element of $\{1,2,\ldots,N\}^H$. For each behavior in a given sequence, we can use the corresponding dynamics model to predict what the state would be after executing that behavior. We are effectively looking $H \times L$ timesteps forward, since each dynamics model predicts $L$ steps forward. We want to minimize the distance from the predicted resulting state and target state:

\begin{equation}
\min_{b_1,...,b_H} \| g^m - f^{b_H}(...f^{b_2}(f^{b_1}(s_t^m))) \|_2^2
\end{equation}
where $s^m_t$ is the current state (from which the predicted rollout begins),  $g^m$ is the target state, $b_1,...,b_H$ is a sequence of behaviors, and $f^{b_i}$ is the dynamics model for behavior $\pi_{b_i}$.

To do this optimization, we sample $K$ sequences uniformly from $\{1,2,\ldots,N\}^H$ and select the best sequence according to the above cost function. Then, we command the low-level module to execute a single step of the first behavior in the sequence. We repeat this procedure at every timestep.  If the set of behaviors were too large, we could sample these sequences from a distribution that we update at every timestep, as is done in Model Predictive Path Integral (MPPI) control \citep{mppi}.


In contrast to other model-based methods that use MPC, our method only needs to do an optimization over a discrete action space, in particular an action space with cardinality equal to the number of behaviors. Because of this, we can use significantly fewer samples $K$ which allows for better real-time performance. If we had learned the dynamics model on the environment directly i.e. $s_{t+1} = f(s_t, a_t)$, then the MPC would need to optimize over the continuous, possibly high-dimensional, action space of the environment \citep{nagabandi}. The same would be true if we used a continuous behavior space \citep{dads}.

Intuitively, the low-level module provides short-horizon model-free behaviors that can be composed by the mid-level module, through model-based learning, to achieve medium-horizon goals. We do not discount that certain environments may benefit from having a hierarchy of model-free layers. However, when the set of model-free policies become sufficiently predictable and useful, a model-based method can exploit this and yield sample-efficiency improvements and add flexibility. Benchmarking has shown that in environments where dynamics are predictable, model-based methods can be more sample-efficient than model-free methods \citep{benchmarkmodelbased}, as one would intuitively expect.



\subsection{Top-level: Planning System}
\label{sec:toplevel}

The top-level module handles planning. The planning system learns a graph that captures the structure and connectivity of the environment. The agent can then traverse this graph to reach desired states. The graph could be defined on the full state space, but in high-dimensional state spaces like that of the Ant, a graph representation would not scale well and thus, would not be very useful. Instead, the graph is defined on the dimensions of interest.

In general, we can apply graph-based planning in domains where the state space can be discretized. For example, classical planning methods for multi DOF robotic arms construct a graph of the collision-free configurations of the robot arm and directly plan paths on this graph. Importantly however, we must be able to reliably move between adjacent nodes in the graph. For a robotic arm, this is straightforward, given the kinematics. But for the Ant, this is not as clear. In our work, the behaviors and mid-level MPC serve as a way to move between nodes. In a sense, the hierarchy serves as a form of ``dimensionality reduction'' as the top level only needs to consider the state space dimensions that are directly relevant for the target task.

To learn the graph, we first allocate potential nodes arranged in a grid across the dimensions of interest. As the Ant moves around the environment, it associates itself to the closest node in the graph and marks it as visited. When the Ant moves between two nodes, it records this as a feasible edge. And likewise, when it attempts to travel between nodes and fails to do so, it marks an edge as blocked. We use a simple exploration strategy in our experiments, by simply selecting the closest unexplored node outside of our connected component. There has been extensive work in the field of graph theory about online graph exploration, which differs from graph search because we must account for the cost of traversing to the node we want to explore \citep{onlineexploration}.

\subsection{Algorithm Summary}

\begin{wrapfigure}{R}{0.5\textwidth}
\begin{minipage}{0.5\textwidth}
  \begin{algorithm}[H]
    \caption{Given graph $G$, dynamics models $\{f^1,...,f^N\}$, behavior library $\{\pi_1,...,\pi_N\}$, the maximum steps $M$ allowed to reach a subgoal, and the distance threshold $T$ to evaluate success}
    \label{fig:algorithm}
    \begin{algorithmic}
      \State g $\leftarrow$ \Call{SelectGoal}{$s_t, G$}
      \State $p_1,...,p_d$ $\leftarrow$ \Call{PlanPath}{$s_t, g, G$}
      \State $i, c$ $\leftarrow$ $1, 0$
      \While{$i \leq d$}
        \If{$c > M$}
            \Return false
        \EndIf
        
         \State $b_1,...,b_H$
         $\leftarrow$ \Call{MPC}{ $s_t, p_i, \{ f^1,...,f^N \} $}
         \State $s_{t+1}$ $\leftarrow$ State after executing $\pi_{b_1}(s_t)$
         \If{$\text{dist}(s_{t+1}, p_i) < T$}
            \State $i, c$ $\leftarrow$ $i + 1, 0$
        \EndIf
         \State $c$ $\leftarrow$ $c + 1$
         \State $s_t$ $\leftarrow$ $s_{t+1}$
      \EndWhile
      \Return true
    \end{algorithmic}
  \end{algorithm}
\end{minipage}
\end{wrapfigure}

The three-level hierarchy consists of the \textit{behavior library}, \textit{model learner}, and \textit{planning system}. The low-level behavior library consists of a set of policies that operate directly on the environment's continuous state-action space. The mid-level operates on a continuous state space but a discrete action space, since the behavior dynamics and target state use continuous states but the actions are to select one from a fixed number of behaviors. The high-level module operates on a discrete state and discrete action space, since the nodes of the graph are points in the state space and the actions are to select one of the finite number of nearby vertices of the graph to traverse to.

As shown in Algorithm \ref{fig:algorithm}, actions are selected by a top-down procedure in which each module commands the module below, and the low-level module executes the action in the environment. First, the planning system selects a goal node in the graph and plans a path to the goal. For each state in the path, we specify it as the target state for the mid-level module. The MPC uses the learned behavior dynamics to optimize a sequence of behaviors that minimizes the distance to that target state. Then, it commands the low-level module to execute one step of the first behavior in this optimal sequence. The low-level simply evaluates the policy given the current proprioceptive state and executes the action. This is repeated for each state in the path until the goal is reached, and we replan a new path if we have deviated from the desired path. 

\section{Related Work}
\textit{Hierarchical RL: } The use of hierarchy is an extensively studied topic in reinforcement learning \citep{dayan}. The options framework \citep{sutton} has led to a line of work that learns layers of policies, some even being trained completely end-to-end with only the task reward \citep{heess, peng}.   

\textit{Skill Discovery: } Learning a useful set of low-level behaviors is a difficult ability. Mutual information has been used in reinforcement learning to encourage exploration \citep{rezende, hoothrouf} and recent work has applied mutual information based objectives to learn skills, often through use of variational bounds \citep{dads, diayn, gregor, thomas, florensa, achiam, hausman}.  A common theme among these works is learning a ``diverse'' library of skills. \cite{dads} additionally emphasizes that behaviors should be predictable. This is well motivated because, similarly to our work, they seek to compose the learned skills using a model-based meta-controller. While the focus of our work is not on how to learn these skills, our results do point out particular advantages and disadvantages of previously proposed methods (see Section \ref{sec:learnthebehaviorlibrary}). Similarly to our work, many of these works have suggested that we can improve the relevance of learned skills to the target task by splitting the inputs and hiding certain inputs from some modules.


\textit{Model-based RL: } After learning the behaviors, we used model-based RL (MBRL) as the meta-controller, similarly to \cite{dads}. However, in contrast, we use a discrete behavior space and also predict multi-step dynamics rather than one step dynamics. Recent work has applied deep neural networks to learn the environment dynamics \citep{chua, nagabandi, gal, lenz} and use this model to select actions, often using MPC. Another work investigates how to use ensembles of value-functions to capture uncertainty and guide exploration \citep{polo}. Our work uses standard MPC, however uses a graph as a more explicit representation of the environment, which aids in guiding the MPC to effectively explore. 

\textit{Planning + RL: } Recently there has been increased interest in connecting planning and RL. Two similar works explore this in the context of an indoor navigation task \citep{building1,building2}. Both evaluate using a mobile robot. Another two works propose methods for organizing observations using a graphical memory \citep{sptm, sorb}. These works evaluate on visual navigation tasks receiving images as observations. In contrast to all of these works, our architecture must handle both high-dimensional state and complex environment dynamics.

\section{Experiments}

Our experiments are focused on evaluating \textit{sample-efficiency} and \textit{flexibility}, as we see these as the fundamental directions to expand the applicability of RL. We selected a series of tasks in the Mujoco Ant environment \citep{gym}. We evaluate as if we were controlling a real-world robotic Ant. Therefore for our method, we do not allow the Ant to ``teleport'' to a previously visited state or reset to a start state, since this would be non-physical and not realizable with a real world robot. If the Ant falls over, we allow it to reset to a proper orientation, but it must do so in place.

\subsection{Learning the Behavior Library}
\label{sec:learnthebehaviorlibrary}

For all the experiments in the sections below, we trained 4 behaviors, for the cardinal directions in the x-y plane, each with 400K environment steps, for a total of 1.6M environment steps. For comparison, SAC-LSP \citep{tuomas} uses 4M steps, DADS \citep{dads} uses 20M steps, and HRL EP3 uses 32M steps \citep{hrl}. While we do not claim that our behavior learning procedure is better, since here we have not done an extensive enough evaluation across tasks and environments, this comparison does suggest two points: (1) Learning a low-level policy conditioned on a continuous latent parameter (as is done in SAC-LSP and DADS) may be wasteful, in terms of samples, for tasks where only a few behaviors are needed and the continuously varying behavior space is less useful. (2) Learning a set of low-level policies through multiple random initializations (as is done in HRL EP3) may lead to overlapping policies. 

\begin{figure}[t]
\centering
\includegraphics[width=.3\linewidth]{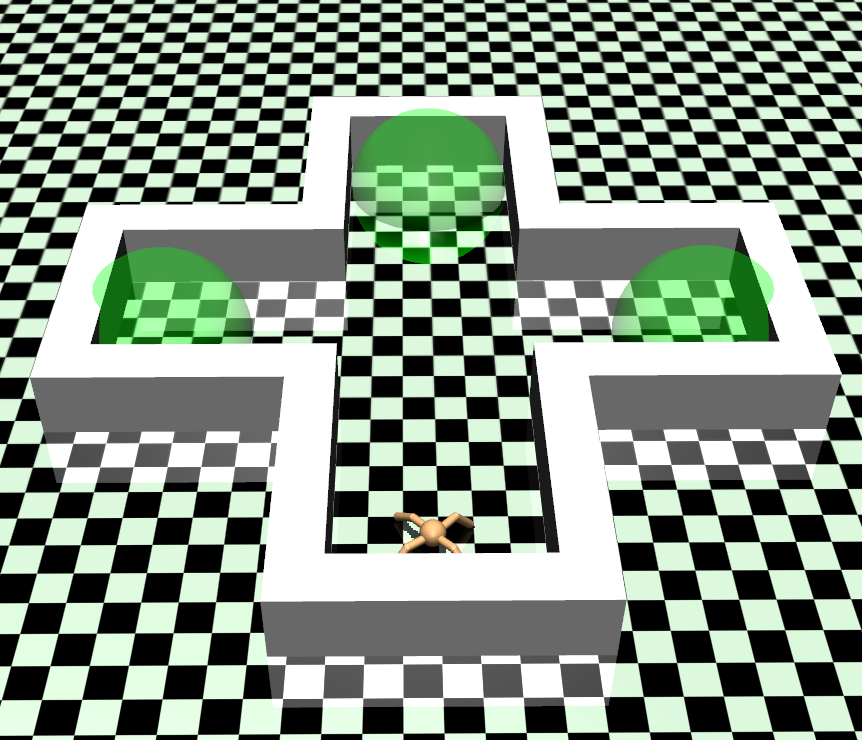}
\hspace{0.1in}
\includegraphics[width=.3\textwidth]{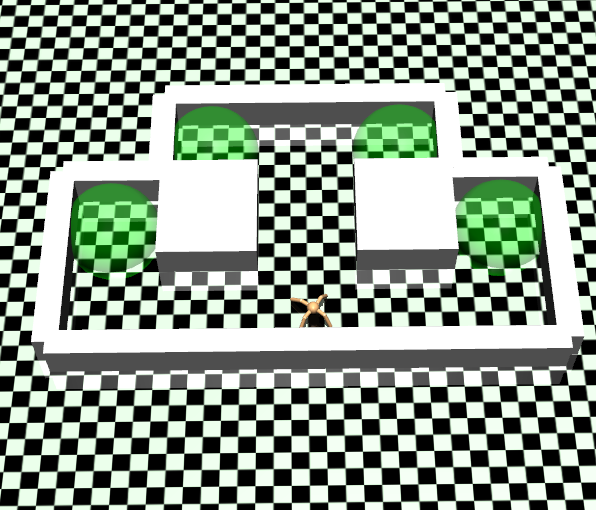}
\hspace{0.1in}
\includegraphics[width=.3\textwidth]{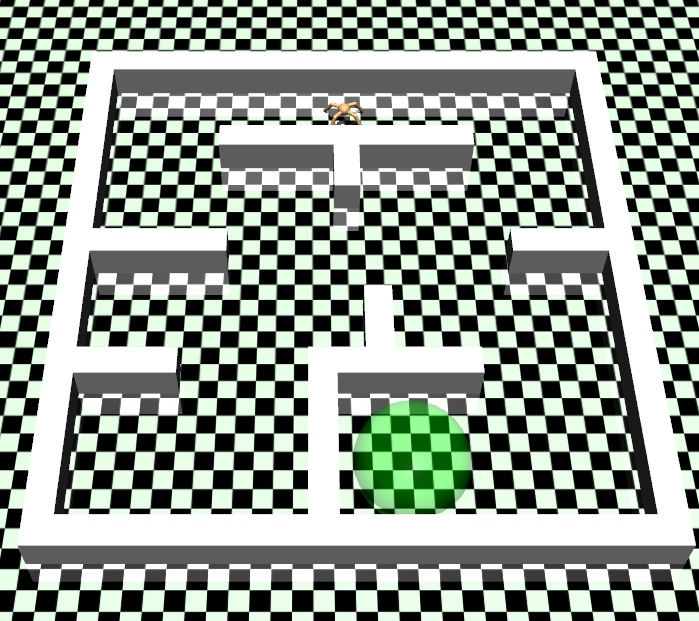}
\caption{\textbf{Maze Environments}: \textit{(Left)} Cross Maze environment with 3 possible goals. \textit{(Center)} Skull Maze environment with 4 possible goals. \textit{(Right)} Complex Maze with a goal revealed at test time.}
\label{fig:mazes}
\end{figure}

\subsection{Free Space Navigation}

We first consider a simple task in which the Ant must reach a specific x-y coordinate, selected randomly from $[-15,15]^2$ each episode. We compare with the results reported in \cite{dads}, which evaluates against state-of-the-art model-based RL \citep{chua}. The metric is the normalized distance to goal which is computed as $\sum_{t=1}^{T} \frac{\|s_t - g\|_2}{T \|g\|_2}$ over a episode length of $T = 200$ timesteps. Our method achieves $0.49 \ (0.17)$, DADS achieves $~0.35$, and the model-based method achieves $~0.60$, as reported in their work. Our method is outperformed by DADS, likely due to the additional samples used to train the low-level policy, however we are able to outperform the model-based method. Note, the model-based method is trained on the specific task of reaching x-y coordinate goals while DADS and our method do not need additional training, though DADS does use significantly more samples to train their low-level policies than our method.

While model-predictive control, through a multi-step optimization, does allow us to consider longer horizons when selecting actions, there are settings in which it will clearly fail due to its optimization of a local objective. For example, in all of the maze environments used in the following experiments, MPC on its own would not be able to properly navigate around obstacles.

\subsection{Waypoint Navigation}

\begin{figure}[t]
\centering
\includegraphics[width=.34\linewidth]{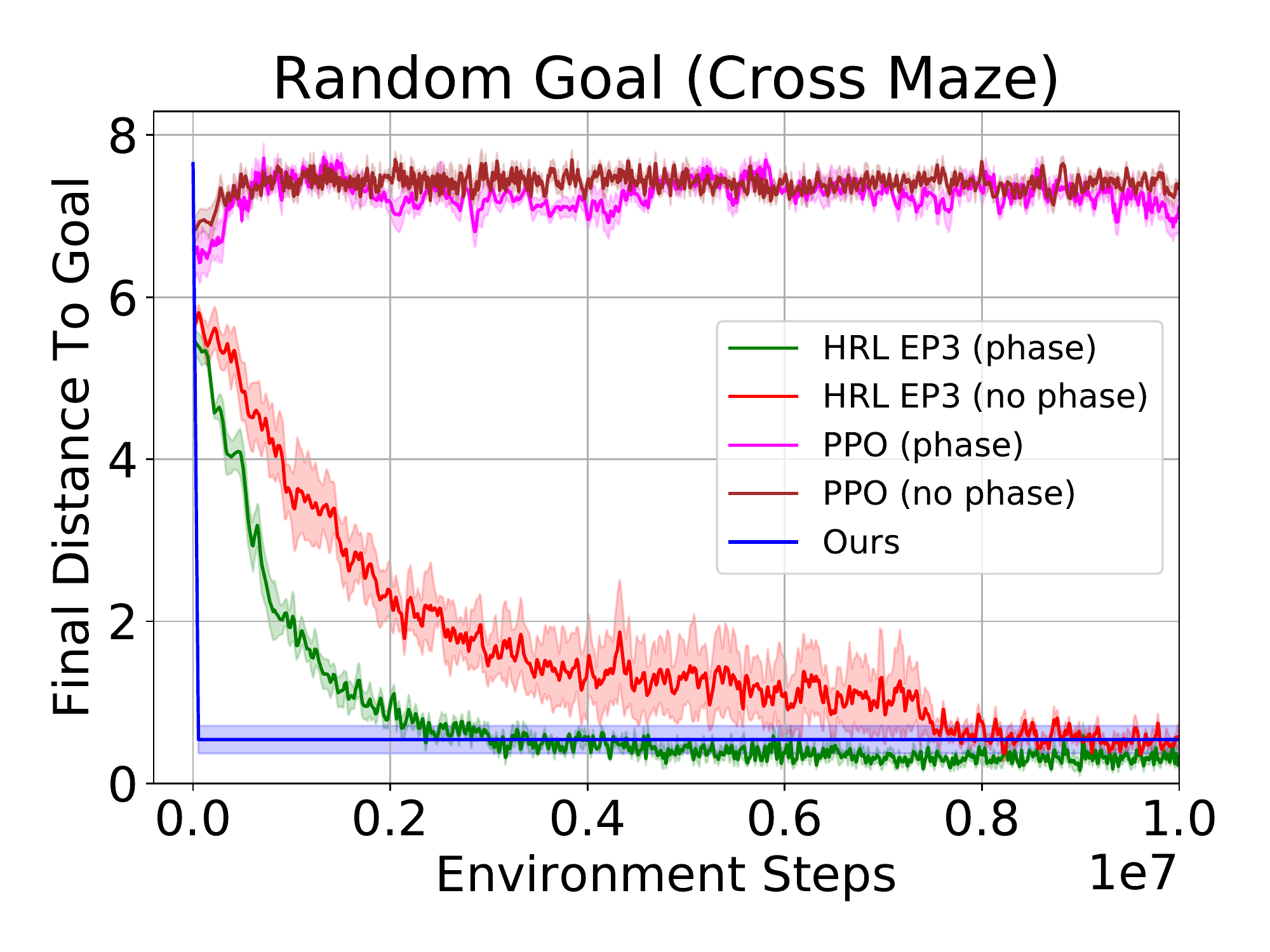}
\includegraphics[width=.34\linewidth]{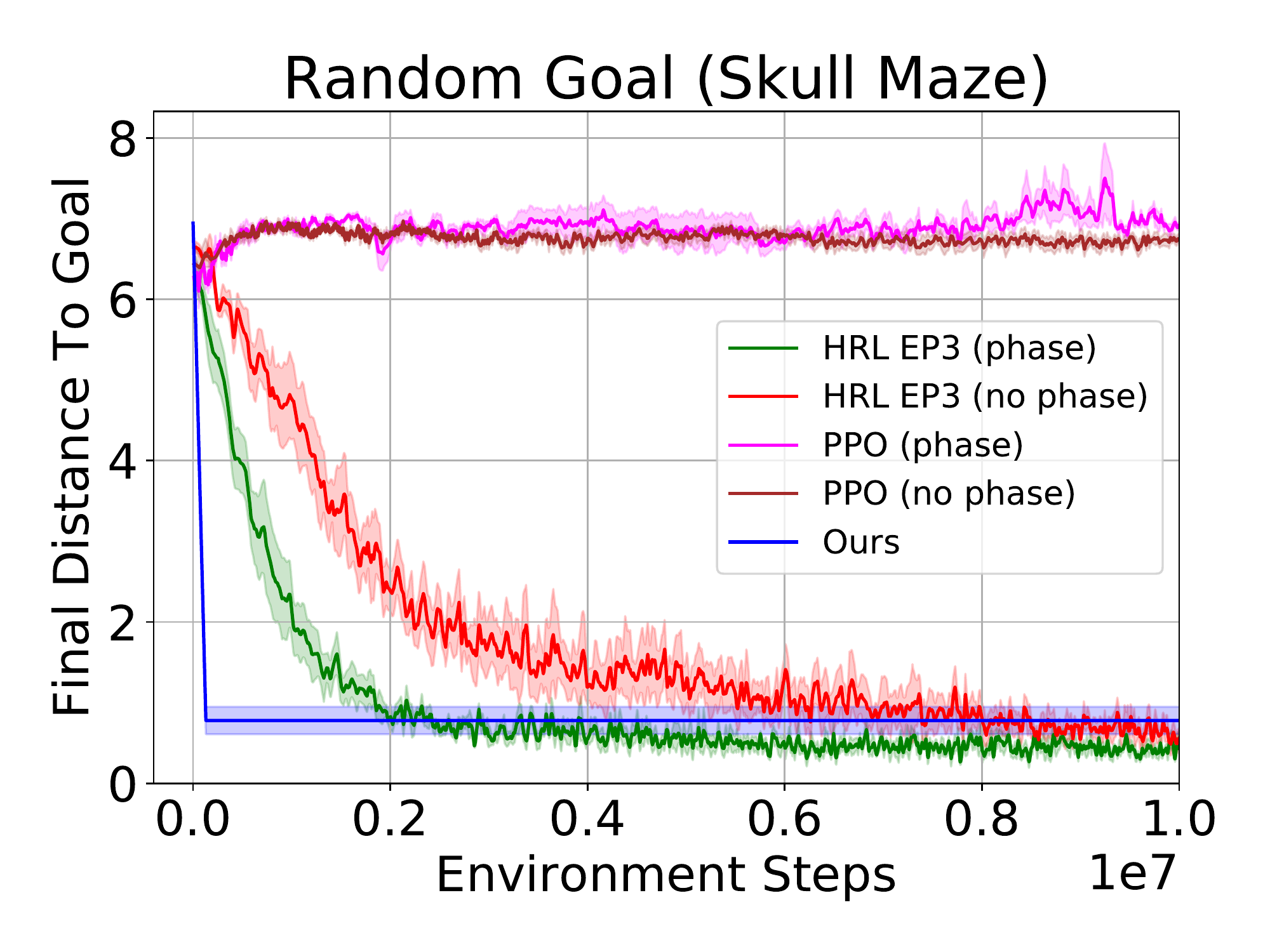}
\includegraphics[width=.3\linewidth]{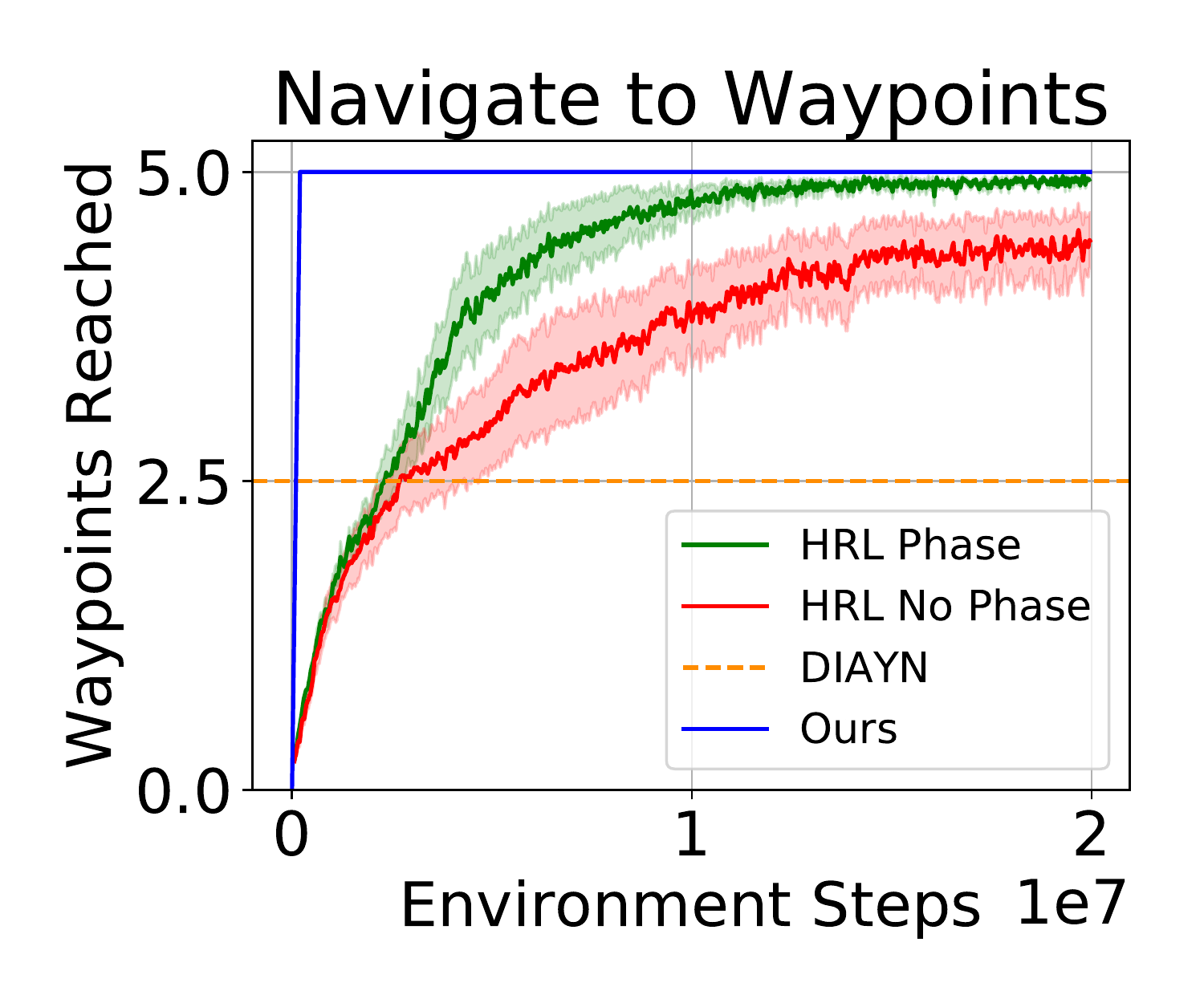}
\caption{\textit{(Left, Center)} \textbf{Random Goal Maze Navigation Performance}: Performance on navigating to a goal randomly selected each episode in the Cross and Skull Mazes. Comparisons are with HRL EP3 and a flat PPO model \citep{ppo}. \textit{(Right)} \textbf{Waypoint Navigation Performance: }  Performance on navigating to a series of waypoints in order. Comparisons are with HRL-EP3. For DIAYN, we report the maximum achieved performance since no evaluation with respect to number of samples was provided, though they reported it required 15 hours of training.}
\label{fig:randommazeperformance}
\end{figure}

We consider a task of navigating to a series of 4 waypoints in order and then returning to the first. This, as well as all the experiments in the following sections, is a sparse reward task where the reward is only received when a waypoint is reached and if it is the next waypoint in the series. In Figure \ref{fig:randommazeperformance}\textit{(Right)}, we compare against the methods in \cite{diayn} and \cite{hrl}. Our method quickly identifies the waypoints with a simple exploration strategy and then can plan a path allowing it to consistently reach all 5 waypoints.

\subsection{Mazes: Fixed Goal}

\begin{wrapfigure}{r}{0.45\textwidth}
\begin{minipage}{0.45\textwidth}
\centering
\includegraphics[width=\linewidth]{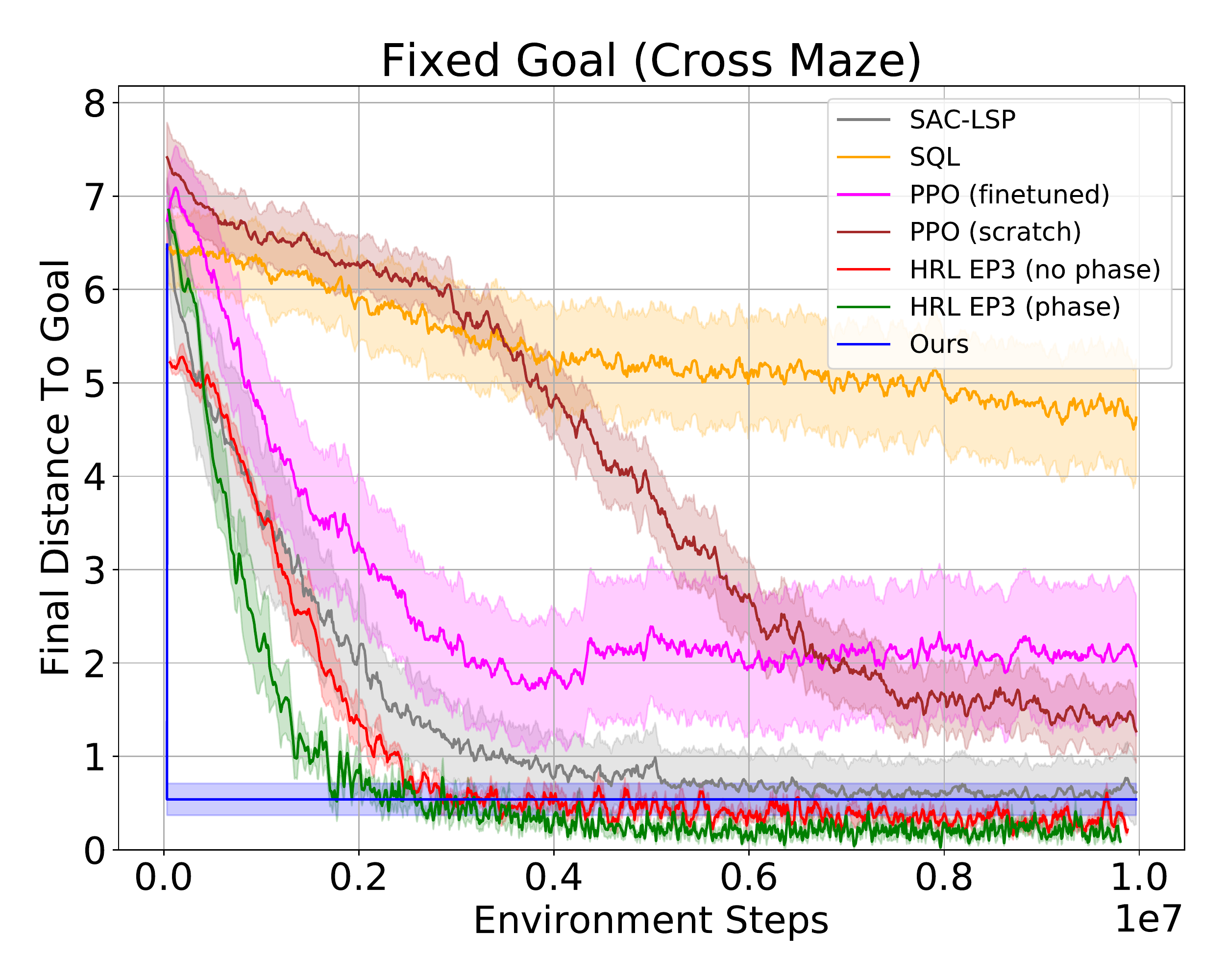}
\caption{\textbf{Fixed Goal Maze Navigation Performance}:  Performance on navigating to a fixed goal in the Cross Maze. The methods compared against include SAC-LSP \citep{tuomas}, soft Q-learning (SQL) \citep{sql}, HRL EP3 \citep{hrl},  and a flat PPO model \citep{ppo}.}
\label{fig:fixedmazeperformance}
\end{minipage}
\end{wrapfigure}

Next, we consider the task of navigating to a fixed target in the simple maze shown in Figure \ref{fig:mazes}\textit{(Left)}. This maze task is introduced by \cite{tuomas} and performance is further improved in \cite{hrl}. Figure \ref{fig:fixedmazeperformance} shows that we learn to do this task in significantly fewer samples than all other methods that are evaluated. Note that we do not include the samples needed to learn the low-level behaviors as these are not included for the other methods either. Nevertheless, as we discussed above, our method even uses fewer samples than the other methods to learn the behaviors. While we found that, at convergence, some of the other methods can more consistently reach the goal, we are more interested in the performance given a limited number of samples. Consider that it takes almost 2M environment steps for next best method (HRL EP3) to match ours. In terms, of wallclock execution time on a real-robot, if each timestep took 0.1s to execute, 2M samples would require $2.3$ days to collect. In contrast, given the values we report for full exploration of the maze in Table \ref{tab:mazes}, our method would take only $35$ minutes.

While this task does demonstrate the sample-efficiency advantages that our method provides, it does not test for the flexible, general-purpose understanding an agent would acquire from exploring a novel environment.

\subsection{Mazes: Random Goals}

\begin{wrapfigure}{R}{0.5\textwidth}
\begin{minipage}{0.5\textwidth}
\centering
\begin{tabular}{lcc}
\toprule
Maze & Explore & Goal \\
\midrule
\textsc{Cross} & 20704.8 (6043.8) & 208.8 (16.9) \\
\textsc{Skull} & 30846.8 (560.7) & 229.6 (37.9) \\
\textsc{Complex} & 95830.6 (15830.6) & 991.8 (152.7) \\
\bottomrule
\end{tabular}
\captionof{table}{\textbf{Environment Steps for Exploration and Goal Reaching :} Averaged over 10 runs of our method.}
\label{tab:mazes}
\end{minipage}
\end{wrapfigure}

We introduce randomness to this task by randomly selecting one of three goal locations in each episode. We also introduce another more difficult Skull maze (shown in Figure \ref{fig:mazes}\textit{(Center)}) which has 4 possible goals. Note, in each episode, the active goal is specified as input to the agent. Figure \ref{fig:randommazeperformance}\textit{(Left, Center)} shows that, again, in both these tasks our method has significant sample-efficiency advantage over other methods. This task is a clear example of how a graph can be a valuable representation. Figure \ref{fig:graphs}\textit{(Right)} shows the graphs learned in the Cross and Skull Maze. They capture the connectivity as well as the borders of the maze. Given this representation and the ability to move between connected nodes, not only is the agent able to move to any of the goal positions, but they can move between any two valid positions in the maze by planning.

\subsection{Complex Maze}

\begin{figure}[t]
\centering
\includegraphics[width=0.6\textwidth]{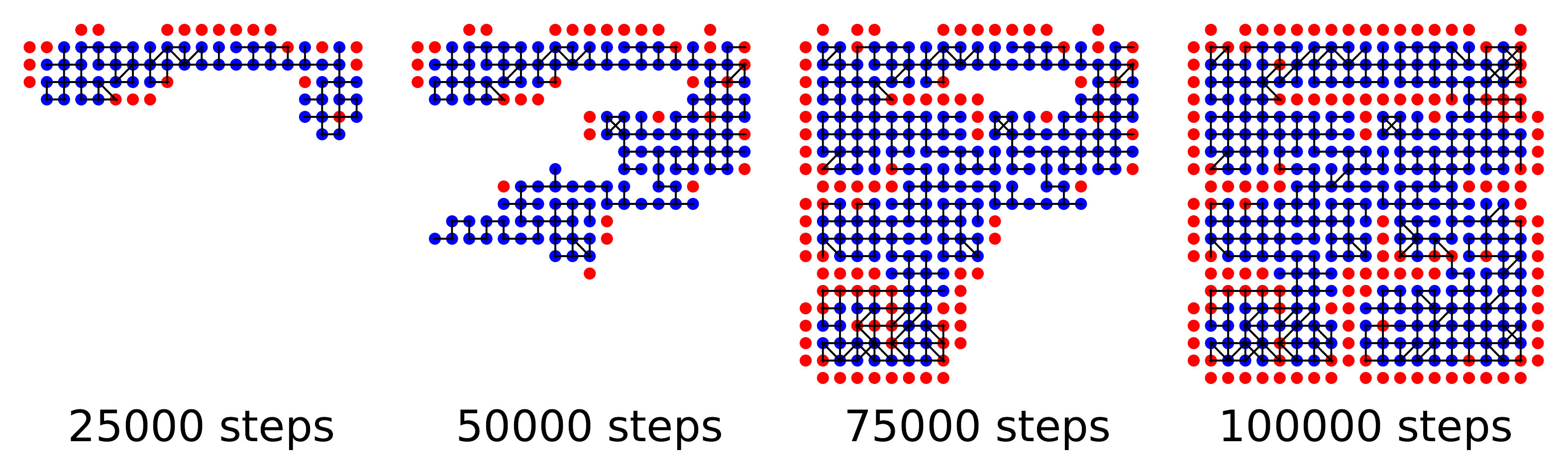}\includegraphics[width=0.4\textwidth]{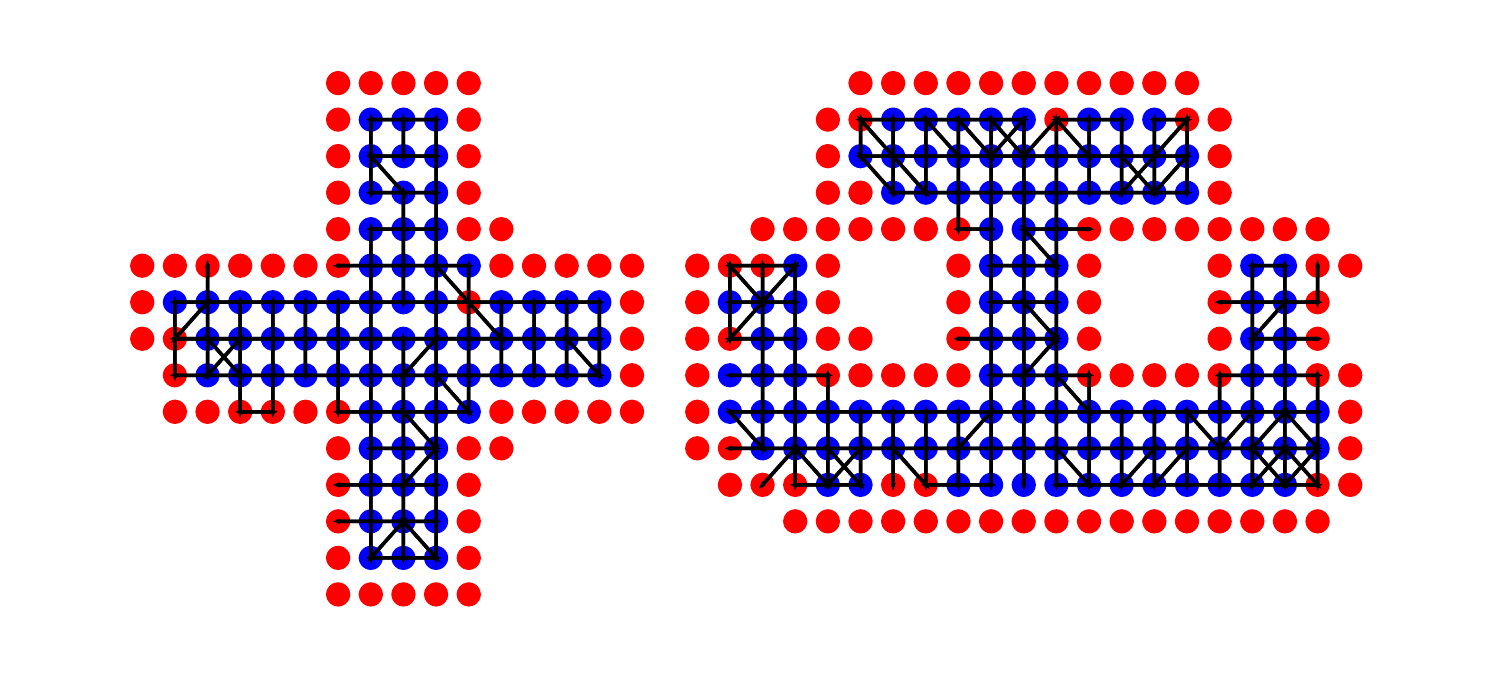}
\caption{\textbf{Learned Graphs: } \textit{(Left)} Evolution of the graph representation during exploration in the Complex Maze. \textit{(Right)} Final learned graphs in the Cross and Skull Mazes.}
\label{fig:graphs}
\end{figure}

Finally, as an illustrative example of the class of problems in which our method excels, we evaluate on a task where an environment is presented at training time (with no rewards) and then at test time, a target state is revealed and the agent must navigate to the target. This tests the agent's ability to efficiently explore and build a model of a novel environment, that it can later use to navigate to any state in the maze. We consider the complex maze in Figure \ref{fig:mazes}\textit{(Right)}. As shown in Figure \ref{fig:graphs}\textit{(Left)}, the agent is able to rapidly explore and build a map of the environment. In Table \ref{tab:mazes}, we report the average number of steps to explore and reach the goal at test time. Standard RL methods will be unable to perform this task, as the policy will need to learn paths to all states in the maze. Our system, which factorizes the task into path planning and path following, can solve it efficiently. We encourage future work in this direction to evaluate performance on complex environments like this, as it will test the limits of standard RL and push us to integrate other representations and algorithms  into the RL framework.

\section{Discussion}

We presented a simple, hierarchical agent architecture that bridges model-free RL, model-based RL, and planning. While the low-level module of our architecture must interact with the continuous state-action space of the environment, through abstraction provided by behaviors combined with MPC, our top-level module can operate on a discrete state-action space. In our experiments, we demonstrated the advantages this provides in terms of sample-efficiency and flexibility. We see three key future directions for our work: (1) Our results have shown that with more training time, a model-free method can eventually be more consistent than our architecture, since it is able to learn and finetune the higher-level policy while our method uses MPC, with no learning. We plan to investigate how to refine the MPC routine with additional experience, possibly by adapting the cost function, in order to improve our method's consistency. (2) In this work, we learn a behavior library independently of other levels of the hierarchy. But it might be the case that the Ant is blocked by an obstacle that it must jump over, and as a result, it must instruct lower-level systems to learn the jumping behavior. Properly selecting behaviors that will actually be useful for the target task will be an important next step in further reducing the total samples required to learn. (3) Finally, we want to push the limits of incorporating planning into RL. Though our work introduces planning into a task that is traditionally solved by fully RL architectures, we are only able to plan at a coarse level. If instead, we could operate on a graph defined at the level of moving specific limbs and joints, we could possibly learn new behaviors and complete new tasks purely through graph search, with no additional training. Humans are capable of this type of zero-shot transfer and in our future work, we hope to demonstrate this is possible for artificial agents as well.



\bibliography{bib}{}

\newpage
\appendix
\section{Appendix}
\subsection{Hyperparameters}

For the low-level \textit{behavior library}, we use Twin Delayed DDPG to learn policy and Q function networks. For the mid-level \textit{model learner}, we used neural networks to model the behavior dynamics and MPC to select behaviors to activate. For the high-level \textit{planning system}, we used a graph representation. In Tables \ref{tab:td3}, \ref{tab:mid}, and \ref{tab:high}, we list the corresponding relevant hyperparameters.

\begin{table}[h!]
\centering
\caption{Behavior Learning Parameters}
\label{tab:td3}
\begin{tabular}{ll}
\toprule
\textbf{Parameter} & \textbf{Value} \\
\midrule
Num. Initial Steps of Random Actions & $1e5$ \\
Num. Initial Steps before Learning & $1000$ \\
Steps between Parameter Updates & $50$ \\
Maximum Episode Length & $1000$ \\
Replay Buffer Size & $1e6$ \\
Batch Size & $100$ \\
Policy Update Delay & $2$ \\
Gamma (Discount Factor) & $0.99$ \\
Polyak Interpolation Factor & $0.995$ \\
Policy Learning Rate & $1e-3$\\
Q Function Learning Rate & $1e-3$\\
Target Policy Noise (std. dev.) & $0.2$\\
Target Policy Clip Threshold & $0.5$\\
Action Noise (std. dev.) & $0.1$\\
Hidden Layer Sizes & [$256$, $256$] \\
Hidden Layer Activation & ReLU \\
Policy Network Output Activation & Tanh \\
Q Network Output Activation & Identity \\
\bottomrule
\end{tabular}
\end{table}

\begin{table}[h!]
\centering
\caption{Model Learning Parameters}
\label{tab:mid}
\begin{tabular}{ll}
\toprule
\textbf{Parameter} & \textbf{Value} \\
\midrule
Hidden Layer Sizes & [$256$, $256$] \\
Hidden Layer Activation & ReLU \\
Output Activation & Identity \\
$L$ (Prediction Time Scale) & $3$ \\
$H$ (MPC Horizon) & $2$ \\
MPC Behavior Prediction Steps  & $2$ \\
$K$ (Behavior Sequence Samples) & $16$ \\
\bottomrule
\end{tabular}
\end{table}

\begin{table}[h!]
\centering
\caption{Graph Learning Parameters}
\label{tab:high}
\begin{tabular}{ll}
\toprule
\textbf{Parameter} & \textbf{Value} \\
\midrule
Grid Interval along x axis & $1.0$ \\
Grid Interval along y axis & $1.0$ \\
$M$ (Maximum steps allowed to reach each subgoal) & $100$ \\
$T$ (Distance threshold to evaluate success) & $0.5$ \\
\bottomrule
\end{tabular}
\end{table}

\subsection{Environment Parameters}

In the Waypoint Navigation task, we use reduced gear ratios of $30$ since this is what is used by the methods we compare with. We also use this reduced ratio for the Complex Maze task. In all other evaluations, we use the standard gear ratios of $150$ to match the methods we compare with.

\end{document}